\documentclass{article}
\usepackage{arxiv}
\usepackage{amsfonts} 
\usepackage{graphicx}
\usepackage{amsmath}
\usepackage{url} 
\usepackage{booktabs}
\usepackage[section]{placeins}
\usepackage{bm}
\usepackage{svg}
\usepackage[export]{adjustbox}
\usepackage{booktabs}
\usepackage{caption}

\makeatletter
\newcommand{\printfnsymbol}[1]{%
  \textsuperscript{\@fnsymbol{#1}}%
}

\usepackage{pifont}
\usepackage{soul}
\newcommand{\cmark}{\ding{51}}%
\newcommand{\xmark}{\ding{55}}

\definecolor{ForestGreen}{RGB}{5,166,88}
\definecolor{LavaRed}{RGB}{222,48,28}
\definecolor{LightGrey}{RGB}{180,180,180}

\begin{document}

\title{Paste, Inpaint and Harmonize via Denoising:  Subject-Driven Image Editing with Pre-Trained Diffusion Model}
\author{
\textbf{Xin Zhang}\thanks{Equal Contribution}  \quad \textbf{Jiaxian Guo}{\printfnsymbol{1} }   \quad \textbf{Paul Yoo} \quad \\  \textbf{ Yutaka Matsuo}  \quad  \textbf{Yusuke Iwasawa} \\
{The University of Tokyo} \\
\texttt{\{xin, jiaxian.guo, paulyoo, matsuo, iwasawa\}@weblab.t.u-tokyo.ac.jp}\\
\url{https://sites.google.com/view/phd-demo-page}}

\onecolumn{%
	\renewcommand\twocolumn[1][]{#1}%
	\maketitle
	\begin{center}
		\centering
		\captionsetup{width=\textwidth}
		\vskip -6.5mm
\includegraphics[width=\linewidth]{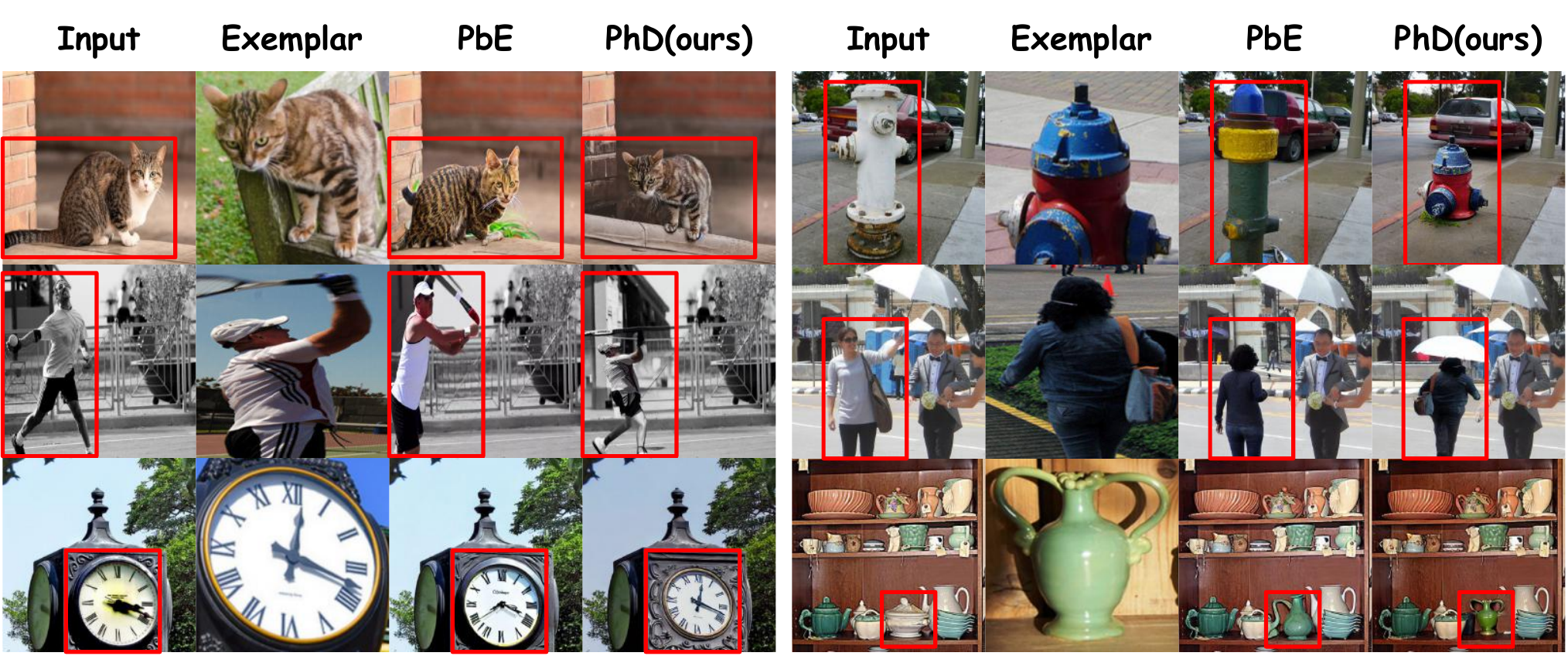}
		\vskip -2mm
		\captionof{figure}{Qualitative comparisons with previous subject-driven image editing methods, where PbE denotes Paint-by-Example~\cite{yang2022paint}. The area outlined by red line denotes the editing area.}\label{fig:front-page}
	\end{center}%
}

\begin{abstract}
Text-to-image generative models have attracted rising attention for flexible image editing via user-specified descriptions.  
However, text descriptions alone are not enough to elaborate the details of subjects, often compromising the subjects' identity or requiring additional per-subject fine-tuning. 
We introduce a new framework called \textit{Paste, Inpaint and Harmonize via Denoising} (PhD), which leverages an exemplar image in addition to text descriptions to specify user intentions.
In the pasting step, an off-the-shelf segmentation model is employed to identify a user-specified subject within an exemplar image which is subsequently inserted into a background image to serve as an initialization capturing both scene context and subject identity in one.
To guarantee the visual coherence of the generated or edited image, we introduce an inpainting and harmonizing module to guide the pre-trained diffusion model to seamlessly blend the inserted subject into the scene naturally. As we keep the  pre-trained diffusion model frozen, we preserve its strong image synthesis ability and text-driven ability, thus achieving high-quality results and flexible editing with diverse texts. 
In our experiments, we apply PhD to both subject-driven image editing tasks and explore text-driven scene generation given a reference subject. 
Both quantitative and qualitative comparisons with baseline methods demonstrate that our approach achieves state-of-the-art performance in both tasks. 
More qualitative results can be found at \url{https://sites.google.com/view/phd-demo-page}.

\end{abstract}

\vspace{-0.2em}
\section{Introduction}



Subject-driven image editing \cite{ruiz2022dreambooth,richardson2023texture,gal2022textual,gal2023designing,kumari2022customdiffusion} is an emerging field that seeks to create realistic, high-quality images by blending user-specified subjects into existing scenes, with potential applications in areas such as photo editing, personalization, entertainment, and gaming. 
For example, one might want to see their pets appearing in a particular movie scene or historical setting. 
%
%
Recent advances in diffusion models have led to increased interest in text-to-image synthesis \cite{ho2020denoising,song2020denoising,rombach2022high,saharia2022photorealistic,ramesh2022hierarchical}. Large-scale diffusion models, trained on vast amounts of text-image pair data \cite{schuhmann2022laion,radford2021learning}, excel at generating high-quality and diverse images based on provided textual descriptions \cite{ramesh2022hierarchical,saharia2022photorealistic}. However, because of the limited expressiveness of text, \emph{i.e.}, even with detailed descriptions, it remains difficult to accurately portray the appearance of a user-specific subject, these models still face challenges in subject-driven image editing.

Going beyond text instruction, Paint by Example (PbE)~\cite{yang2022paint}  proposes a subject-driven editing approach, which enables users to transfer a subject from an exemplar image to a scene image, as illustrated in Figure \ref{fig:front-page}. 
PbE utilizes CLIP \cite{radford2021learning} image embedding to extract subject information from the exemplar image in place of the textual embedding as the condition for the pre-trained diffusion model. By fine-tuning the diffusion model on a self-constructed dataset, PbE is capable of performing subject-driven editing tasks. Despite its potential, Paint by Example faces several challenges:
1) Due to the information extraction bottleneck of CLIP, PbE, while capable of preserving the subject's semantics, compromises low-level details. This can result in distortion of the subject's identity. As shown in Figure \ref{fig:front-page}, PbE can recognize the subject's class but often generates inaccurate details, such as the color of a hydrant and the texture of the vase. 2) Since PbE replaces the textual embedding with that of image embedding from CLIP, the diffusion model loses its text-driven generation capabilities.  This limitation hinders the model's ability to leverage textual guidance for fine-grained control over the generated output. 3) PbE fine-tunes pre-trained diffusion models~\cite{rombach2022high}, which may hinder the original creativity of these models~\cite{mou2023t2i, zhang2023adding}. These problems limit the flexibility of Paint-by-Example when applying it to real applications, 

In this paper, we present \textit{Paste, Inpaint and Harmonize via Denoising} (PhD). 
Unlike PbE, PhD does not require fine-tuning the pre-trained diffusion models and thus does not hinder its original scene generation and text-driven control ability. In addition, PhD does not convert exemplar images into text embedding to prevent potential information loss. 
Instead, PhD comprises two steps: the \textit{Paste Step} and the \textit{Inpaint and Harmonize via Denoising} step. 
In the \textit{Paste Step}, instead of converting the subject information into a textual embedding~\cite{gal2022textual,gal2023designing,ruiz2022dreambooth}, we first utilize an off-the-shelf segmentation model to extract the subject from a user-provided exemplar image, and then directly paste the subject onto a background scene. This way, we encompass both the scene context and the subject into a single image, avoiding the need to fuse the concepts via time-consuming fine-tuning to learn the subject identity.

However, pasting alone does not lead to visually realistic results because, in most cases, the scene context surrounding the subject from the exemplar image and scene image differs significantly, \emph{e.g.}, illumination intensity and direction.
To tackle this issue, we implement the inpainting and harmonizing module. This module aims to blend images in a semantically meaningful manner. Specifically, it takes the pasted image as input and learns to consider both subject and global context information. Drawing inspiration from ControlNet \cite{zhang2023adding}, we incorporate the output of the inpainting and harmonizing module into the feature map of the pre-trained diffusion model~\cite{rombach2021highresolution} rather than replacing the textual embedding. This approach allows us to guide the pre-trained diffusion model in generating context-consistent and photo-realistic images while preserving its text-driven capabilities.

 
In order to train such an inpainting and harmonizing module, we construct a training dataset in a self-supervised manner inspired by the prior work~\cite{yang2022paint}. This involves randomly selecting a subject from an image and applying data augmentations, specifically color distortion, shape change, and rotation, to the chosen subject. The inpainting and harmonizing model is then trained to reconstruct the original image based on the augmented subject, allowing it to learn how to naturally blend the subject into the image according to the surrounding context.
Because the inpaint and harmonize module is the only trainable part in PhD, it offers the following advantages by not training the pre-trained diffusion model's parameters: (1) With fewer learnable parameters, both training time and training data size can be reduced. (2) PhD can preserve the strong image synthesize and concept composition capabilities of the pre-trained diffusion models, thus enabling high-quality image synthesis and text-driven scene generation, as illustrated in Table \ref{tab:comparison}.

\begin{table}[]
\setlength{\tabcolsep}{1.2mm}
\caption{Comparison with prior methods. The columns indicate each method. 1) denotes that editing the existing image with the given subject, 2) denotes that generates the background scene with the given subject, 3) denotes performing editing or generation with a novel subject without training and 4) denotes generating or editing images based on the textual prompt. }
\centering
\begin{tabular}{lccccc}
\hline
  & {S-D \cite{rombach2021highresolution}}  & {PbE \cite{yang2022paint}} & {DCCF \cite{xue2022dccf}} &  {B-D \cite{avrahami2022blended}}&{Ours} \\
\hline
1) Subject-Driven Editing & \textcolor{LavaRed}{\xmark} & \textcolor{ForestGreen}{\cmark} & \textcolor{ForestGreen}{\cmark} & \textcolor{LavaRed}{\xmark} & \textcolor{ForestGreen}{\cmark} \\
2) Scene Generation & \textcolor{LavaRed}{\xmark} & \textcolor{LavaRed}{\xmark} & \textcolor{LavaRed}{\xmark} & \textcolor{LavaRed}{\xmark} & \textcolor{ForestGreen}{\cmark} \\
3) Train-Free for Novel Subject&  \textcolor{LavaRed}{\xmark} & \textcolor{ForestGreen}{\cmark} &   \textcolor{ForestGreen}{\cmark} & \textcolor{LavaRed}{\xmark} &  \textcolor{ForestGreen}{\cmark}\\
4) Text Control &  \textcolor{ForestGreen}{\cmark} & \textcolor{LavaRed}{\xmark} & \textcolor{LavaRed}{\xmark}  & \textcolor{ForestGreen}{\cmark} & \textcolor{ForestGreen}{\cmark} \\
\hline
\end{tabular}

\label{tab:comparison}
\vspace{-1.7em}
\end{table}

In our experiments, we implemented our framework to conduct subject-driven editing tasks on the COCOEE dataset \cite{yang2022paint}. We conduct a variety of evaluations to qualitatively and quantitatively assess the effectiveness of our proposed approach and compare it to existing methods in multiple aspects, including visual quality, subject identity preservation, and semantics preservation. The result demonstrates that our framework surpasses the baselines in most evaluation metrics, especially for visual quality and subject identity preservation.
Moreover, our framework is capable of subject-driven scene-generation tasks using only a single exemplar image featuring the subject. By incorporating textual input, we can further enhance the versatility of subject-driven image synthesis tasks, \emph{e.g.} style transfer. This offers a broader range of possibilities for generating realistic and contextually appropriate images in a variety of subject-driven scenarios.

\vspace{-0.2em}

\vspace{-0.2em}
\section{Related Work}
\subsection{Text-Guided Image Synthesis}
Text-Guided Image Synthesis aims to generate high-quality images based on given textual descriptions. 
Early methods~\cite{tao2020df,xu2018attngan,ye2021improving,zhang2021cross,zhu2016generative,karras2017progressive,karras2021alias,karras2020analyzing} attempted to train Generative Adversarial Networks (GANs)\cite{goodfellow2014generative} on large-scale image-text data. 
However, GAN-based methods suffer from training instability\cite{brock2018large} and mode collapse issues~\cite{srivastava2017veegan}. 
Subsequently, approaches like~\cite{gafni2022make,ramesh2021zero,yu2022scaling,esser2021taming,van2017neural} treated images as sequences of discrete tokens and stabilized the training process.
Recently, works such as~\cite{avrahami2022blended,kim2022diffusionclip,ho2020denoising,song2020denoising,rombach2021highresolution,saharia2022photorealistic,saharia2022palette} have exploited the strong modeling capabilities of diffusion models, achieving superior performance over GANs and demonstrating unprecedented diversity and fidelity~\cite{nichol2021improved,ramesh2022hierarchical,saharia2022photorealistic,rombach2021highresolution}. 
Our model aims to harness the robust image synthesis abilities of pre-trained text-to-image models for subject-driven image editing tasks.
\vspace{-1em}



\subsection{Subject-Driven Image Editing and Generation}

Subject-Driven Image Editing focuses on incorporating specific subjects into a given scene.
A growing body of research~\cite{alaluf2022hyperstyle,bau2020semantic,richardson2021encoding,shen2020interpreting,shen2021closed,nitzan2022mystyle,zhang2020deep} investigates the optimization and interpolation of GAN latent codes to generate images featuring particular subjects. 
In the context of diffusion models, \cite{avrahami2022blended} enables image editing based on subjects from reference images. 
Textual Inversion~\cite{gal2022textual} refines textual embeddings to represent new subjects, while Dreambooth~\cite{ruiz2022dreambooth} fine-tunes pretrained models using concept images. 
CustomDiffusion~\cite{kumari2022customdiffusion} combines the benefits of Textual Inversion and Dreambooth by learning new textual embeddings and fine-tuning cross-attention to capture multiple concepts simultaneously. 
Recent studies~\cite{Gal2023DesigningAE} suggest developing specialized encoders within specific domains to enhance concept learning efficiency. 
Unlike these methods, Paint-by-Example~\cite{yang2022paint} performs subject-driven image editing by extracting subject information using CLIP and fine-tuning pretrained diffusion models, which restricts its flexibility. 
In contrast, our method streamlines the process by eliminating the need to learn specific objects during inference and can perform both subject-driven image editing and generation tasks.  The concurrent work \cite{kulal2023putting} emphasizes placing the same person in various scenes, while our method is capable of positioning not only humans but also other subjects in diverse scenes.
\vspace{-1em}
\subsection{Image Inpainting}
Early works on image inpainting focused on local appearance diffusion methods~\cite{935036,bertalmio2000image,dhariwal2021diffusion} and patch matching techniques~\cite{Barnes:2009:PAR,efros1999texture}. 
More recent approaches leverage larger datasets for feature matching~\cite{hays2007scene,pathak2016context}. 
A learning-based approach was introduced by \cite{pathak2016context}, which inspired subsequent works using CNNs~\cite{liu2018image,sohl2015deep,yu2018generative,yu2019free,zhao2021large,zheng2022cm} and Transformers~\cite{bar2022visual,esser2021taming,yu2021diverse}. 
GANs are also widely used for inpainting, but often require auxiliary objectives related to structures, context, edges, contours, and hand-engineered features~\cite{iizuka2017globally,song2018spg,nazeri2019edgeconnect,yu2018generative,wang2018image}. 
Recently, diffusion models have gained popularity in inpainting tasks~\cite{ho2020denoising,song2020score,lugmayr2022repaint,saharia2022palette}. 
Notably, some works~\cite{nichol2021glide,rombach2022high,avrahami2022blended} have explored guided inpainting using diffusion-based models informed by text. 
PbE~\cite{yang2022paint} uses an exemplar image of the target object instead of text, while \cite{kulal2023affordance} employs an image of a person to be inserted into the scene as an alternative to text. 
In comparison to these methods, this paper aims to preserve low-level features from the exemplar while maintaining the controllability offered by text guidance.
\vspace{-1em}
\subsection{Image Harmonization}
In photo editing, creating realistic composite images often involves cutting the foreground from one picture and pasting it onto another. 
Deep learning methods for color harmonization treat it as an image-to-image translation task, using techniques such as encoder-decoder U-net structures~\cite{tsai2017deep}, learnable alpha-blending masks~\cite{sofiiuk2021foreground}, domain transfer~\cite{BargainNet2021,cong2020dovenet}, attention mechanisms~\cite{cun2019improving,hao2020image}, and generative adversarial networks~\cite{chen2019toward}.
Recent works have focused on high-resolution images using self-supervised learning~\cite{jiang2021ssh}, global parameters~\cite{hu2018exposure,jia2006drag}, and pixel-wise curves to adjust image attributes such as lightness and saturation~\cite{guo2020zero}. 
The recent DCCF~\cite{xue2022dccf} method introduces four novel neural filters and a simple handler for user cooperation, achieving state-of-the-art performance on the color harmonization benchmark. 
However, these methods presume a harmonious relationship between foreground and background, only adjusting low-level color space and keeping the structure intact. 
This paper tackles semantic image composition under challenging semantic disharmony.

\vspace{-0.2em}

\vspace{-0.2em}
\section{Preliminary}
\textbf{Diffusion model} is a family of generative models that can sample realistic images from Gaussian noise by progressively reducing noise \cite{sohl2015deep,song2020denoising,nichol2021improved,ho2020denoising}. They consist of two processes: diffusion and reverse. In the diffusion process, an image $x_0$ is gradually corrupted by injecting random noise of increasing variance through $T$ timesteps with noise level schedules $\{\beta_t\}_{t=1}^T$. A noisy variable $x_t$ at timestep $t$ is constructed via the following equation:
\begin{equation}
    x_{t} = \sqrt{\Bar{\alpha_{t}}}x_{0} + \sqrt{1-\Bar{\alpha_{t}}}\epsilon
\end{equation}
where $\epsilon \sim \mathcal{N}(0,\,1)$, $\alpha_t = 1 - \beta_t$, and $\Bar{\alpha_t} = \prod_{s=1}^t \alpha_s$.

In the reverse process, noisy image $x_{t+1}$ is denoised by estimating the noise $\epsilon$ with a neural network $\epsilon_{\bm{\theta}}(x_{t+1}, t)$. The network parameters $\theta$ are trained by minimizing the $\ell_2$ loss:
\begin{equation}
    L = E_{x_{0}, t, \epsilon}|| \epsilon_{\bm{\theta}}(x_{t+1},t) - \epsilon||^{2}_{2}
\end{equation}

\noindent\textbf{Conditional diffusion model} samples an image from a conditional distribution $x_0 \sim p_{\bm{\theta}}(x \vert \bm{c})$ given conditioning information $\bm{c}$ such as a class label, an image, or text prompt. The resulting noise estimation network can be formulated as $\epsilon_{\bm{\theta}}(x_{t+1}, t, \bm{c})$.
In classifier-free guidance\cite{ho2022classifier}, a linear combination of conditional and unconditional estimates is computed to enhance sample quality at the cost of diversity.
\begin{equation}
    \Tilde{\epsilon} = (1 + \textit{w}) \epsilon_{\bm{\theta}}(x_{t+1}, t, \bm{c}) - \textit{w} \epsilon_{\bm{\theta}}(x_{t+1}, t)
\end{equation}





\section{Method}

\begin{figure*}[h]
    \centering
    \includegraphics[width=\textwidth]{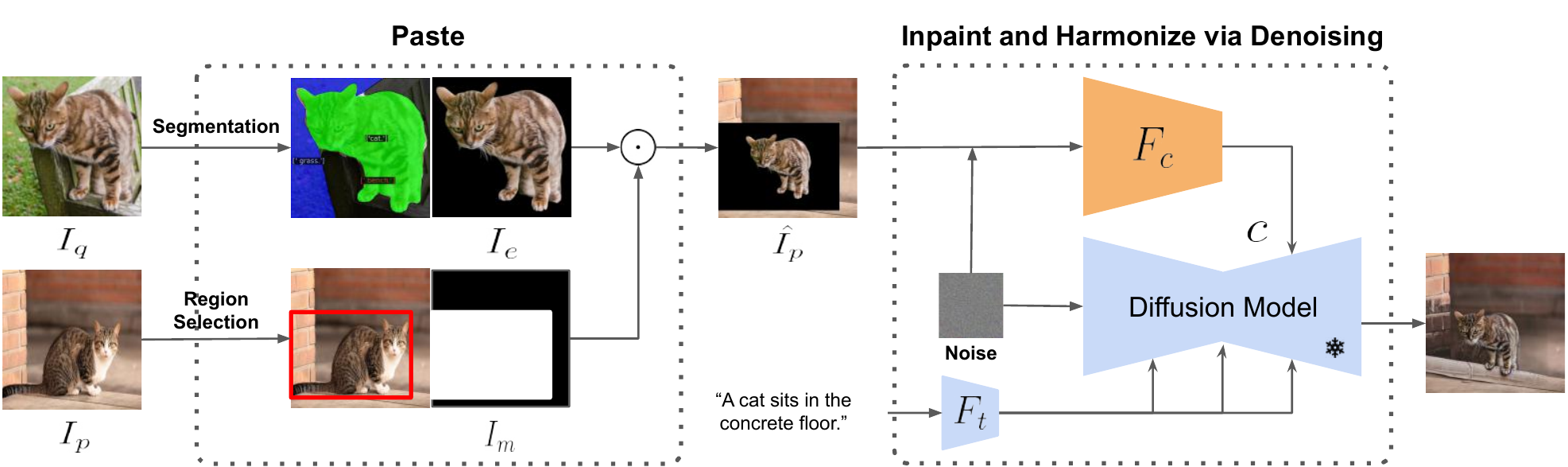}
    \caption{
    The illustration of our proposed Paste, Inpaint and Harmonize via Denoising~(PhD) framework.
    In the Paste step, we extract the subject from exemplar image $I_q$ using segmentation model and removed the background of the objects within the mask $I_m$ to obtain $I_{e}$. 
    Then we paste $I_{e}$ onto the masked scene image to obtain the pasted image $\hat{I_p}$.
    In the  Inpaint and Harmonize via Denoising step, the Inpainting and Harmonizing module $F_c$ takes $\hat{I_p}$ as the input, and output editing information $c$ to guide the frozen pre-trained diffusion models. The text encoder $F_t$ takes textual prompts as input so that it is able to adjust the style or scene in the edited image.
    }
    \label{fig:pipeline}
\end{figure*}

Given an exemplar image $I_q$, users expect to generate or edit the input image $I_q$ referring to the subjects in the exemplar image. 
We propose a two-step processing framework called PhD, comprising the \textit{\textbf{P}aste}, Inpaint and \textit{\textbf{H}armonize via \textbf{D}enoising} steps. As illustrated in Figure \ref{fig:pipeline}, we first extract the subject from the exemplar image $I_q$ and seamlessly paste it into the scene image $I_p$ during the \textit{Paste} step. Subsequently, we utilize a pre-trained diffusion model in the \textit{{H}armonize via Denoising} step to render the edited image photorealistic. In the following sections, we will elaborate on the components of our PhD framework.




\vspace{-0.2em}
\subsection{Paste step}

Given an exemplar image, $I_q$, and a scene image $I_p$, users first select an editing area within the scene image $I_p$. 
For instance, when editing a family courtyard image, users can flexibly select a bounding box in the left or right side of the yard as the editing area. 
Simultaneously, we extract the user-specific subject from the exemplar image $I_q$ using a pre-trained segmentation module, such as U²Net~\cite{Qin_2020_PR} or SAM \cite{kirillov2023segment}.
Upon extracting the subject from the exemplar image, we resize the subject so that its size fits the editing area, and then directly paste it onto the editing area of the background scene image instead of encoding it into an embedding, allowing the subject's details to be directly incorporated into the scene image without losing the information. 
Typically, the editing area's shape is larger than the subject in PhD, so that the inpainting and harmonizing model in the subsequent stage is able to adjust the subject's geometry, direction and even complete the partial subject to match the context of the scene image, as  Figure \ref{fig:pipeline} shows.
For convenience, we denote a binary mask image $I_m$ to distinguish the editing area and the unchanged area within the scene image $I_p$. We denote the extracted object as $I_e$ and the pasted image as $\hat{I_p}$, respectively.





\vspace{-0.2em}
\subsection{Inpaint and Harmonize via Denoising step}

In the \textit{Paste} step, we obtain the editing image, denoted as $\hat{I_p}$, which includes the pasted subject from the exemplar image $I_q$. However, the editing image may appear unrealistic due to differences in context information between $I_p$ and $I_q$, such as illumination intensity and the unreasonable direction of the subject. To solve this, one naive solution in Paint-by-Example \cite{yang2022paint} directly fine-tunes a pre-trained diffusion model to harmonize $\hat{I_p}$. However, this approach struggled to generalize to unseen test images, as altering the parameters in the original pre-trained diffusion models could lead to trivial solutions and the forgetting of concepts learned from large-scale datasets during pre-training as shown in Section \ref{sec:aba}. To prevent this issue, we propose freezing the parameters of the pre-trained diffusion model, preserving its robust image generation and concept composition capabilities \cite{liu2022compositional}. Our goal is to minimize the context gap between $I_p$ and $I_q$, making the edited image $\hat{I_p}$ realistic. We introduce the Inpaint and Harmonize via Denoising step, where an inpainting and harmonizing module is designed to guide the pre-trained diffusion model to inpaint the masked area according to the context information and generate context-consistent images. As shown in Figure \ref{fig:pipeline}, the inpainting and harmonizing module takes the edited image $\hat{I_p}$ and mask information as input and learns how to blend the subject within the mask area into the background.

According to the experimental analysis in  the recent ControNet method \cite{zhang2023adding} and \cite{tumanyan2022plug}, the encoder blocks from U-nets  \cite{ronneberger2015u} are able to extract sufficient structure and texture information from the image. In this way, we adopt the same architecture as the recent ControlNet \cite{zhang2023adding}, which includes the encoder blocks from U-nets with convolution layers in diffusion models, to extract subject details and recognize the relationship between the subject and the background. 
Instead of replacing the textual embedding, we add the extracted information from the painting and harmonizing module into the latent in the pre-trained diffusion model, guiding it to inpaint the masked area and harmonize the subject with the context-consistent images. Specifically, we initialize the  inpainting and harmonizing module using the U-net encoder from the pre-trained diffusion model to leverage its learned information from large-scale training datasets.



This design provides the following advantages: (1) The parameter sizes are smaller compared to the pre-trained diffusion model, which reduces both the training time and the size of the required dataset.  (2) Since the pre-trained diffusion model is non-trainable, we can utilize its strong image generation and concept composition capabilities while avoiding the overfitting issue that may arise in a small training dataset.  Next, we will present how we train our inpainting and harmonizing module.


\begin{figure*}[!htb]
    \centering
    \includegraphics[width=\textwidth]{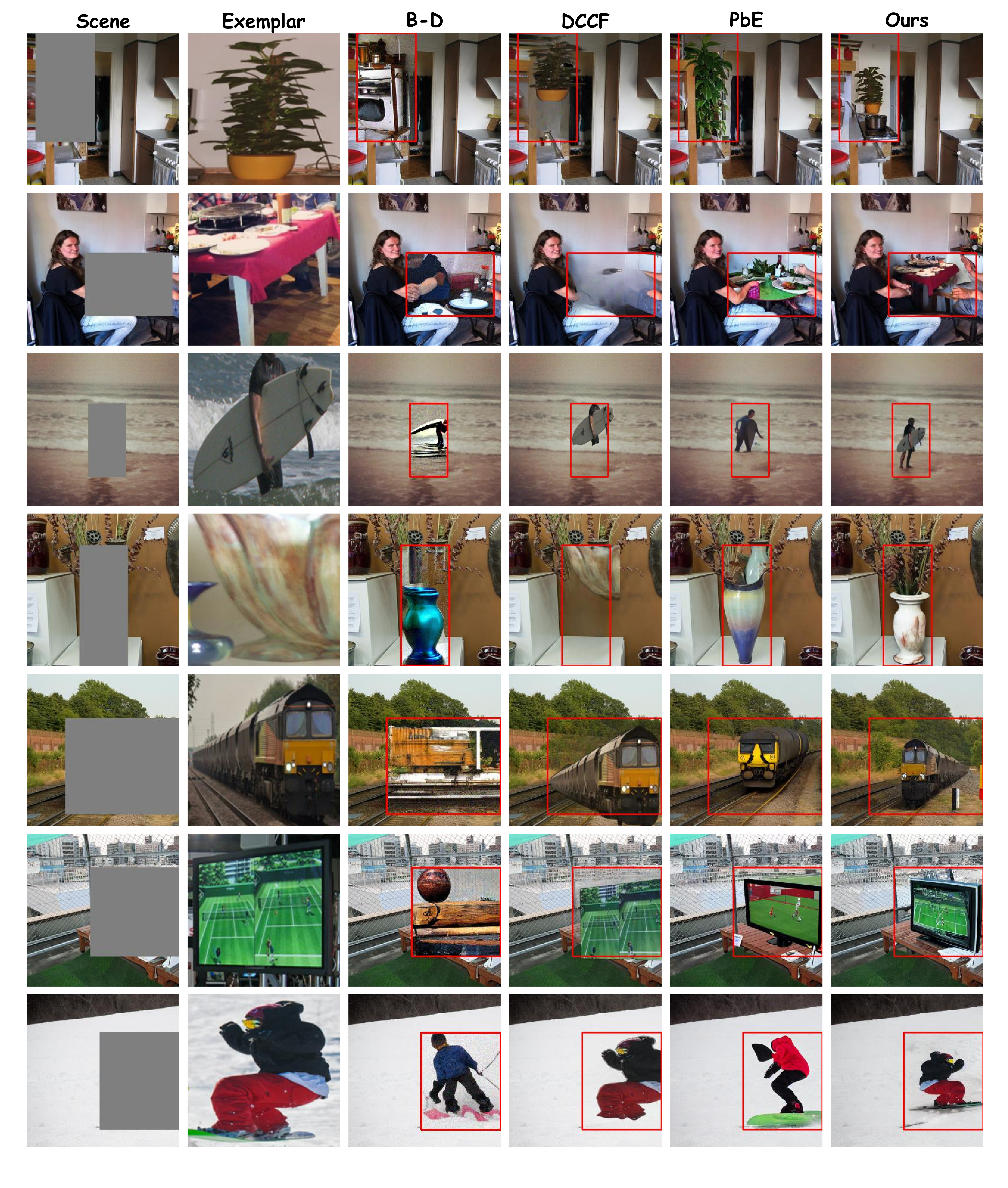}
    \caption{Qualitative results of subject-driven image editing methods, where B-D denotes blended-diffusion model \cite{avrahami2022blended} and PbE denotes Paint-by-Example \cite{yang2022paint}. The results are generated by our method without any further optimization. More quantitative results can be found in Appendix.
        }\vspace{-1em}
    \label{fig:quali_edit}
\end{figure*}

\vspace{-0.2em}
\subsection{Training}
\paragraph{\textbf{Dataset Preparation} }
In order to train our  inpainting and harmonizing module that can flexibly blend any subject into the background scene, it is essential to first construct a training dataset containing paired exemplar objects, input images, and their corresponding edited images.
To construct such a dataset, we use 130k images from the open-images v7 dataset, which comprises 1.9 million photographs reflecting daily life across 600 object categories.
Following the self-supervised approach of PbE~\cite{yang2022paint}, we begin by extracting subjects from the bounding box of the OpenImage dataset~\cite{kuznetsova2020open} and removing the background using the U\textsuperscript{2}Net \cite{Qin_2020_PR} to obtain the subject information.
\vspace{-0.2em}
\paragraph{\textbf{Augmentation Strategy} }To prevent the model from merely copying and pasting, we employ the albumentations library \cite{info11020125} to perform data augmentation on the extracted subjects. Techniques used include HorizontalFlip, Rotate, HueSaturationValue, Blur, and ElasticTransform, each applied with a 10\% probability. By learning how to reconstruct from the image with augmented subjects, the  inpainting and harmonizing module is able to learn how to change the geometry and illustration of subjects to fit them in the background scene.

Subsequently, we add noise into the bounding box to transform it into an irregularly shaped mask with a 50\% probability. This approach makes the mask more closely resemble what users would provide during actual inference. To generate the final edited image, we resize the previously obtained exemplar and place it at the centre of the mask, thereby obtaining the desired edited image for our training dataset. This process ensures a comprehensive and diverse dataset to effectively train our  inpainting and harmonizing module.
\vspace{-0.2em}
\paragraph{\textbf{Training Objective function} }
Our algorithm takes as inputs an edited image $\hat{I_p}$ and an original scene image $I_p$. We first add noise iteratively to $I_p$ to produce a noisy image $I_p^t$. Then, we use the inpainting and harmonizing module $F(\cdot)$ to convert $I_p^t$ into a condition $c$.
Given the time step $t$, text prompts $c_t$, and the extracted subject condition $c$, our goal is to optimize the parameters $\phi$ of the  Inpainting and  Harmonizing module to predict the noise $\epsilon$ that was added to $I_p$ using the following loss function \cite{ho2020denoising,song2020denoising,rombach2021highresolution,zhang2023adding}:
\begin{equation}
\vspace{-0.2em}
     \mathcal{L} = 
     \mathbb{E}_{I_p, t, \epsilon \sim \mathcal{N}(0,\,1)}\left[
     {
     \|
     \epsilon - \epsilon_{\theta}(I_p^t, t, c_t,F_{\phi}(\bm{\hat{I_p}}))
     \|
     }^2
     \right]
     \vspace{-0.2em}
\end{equation}
where $\epsilon_{\theta}$ is the parameter of the pre-trained diffusion model.
\vspace{-0.2em}
\paragraph{\textbf{Training Details}}
We employ the StableDiffusion v1.5 \cite{rombach2022high} as the backbone for our approach and freeze all its parameters. 
To coordinate the images, we utilized ControlNet, which was created by copying the encoder of the unet and served as the condition module. 
During the training process, we set the image resolution as 512x512, the learning rate as 1e-4, and the batch size as 8, We trained the   inpainting and  harmonizing module for 24 hours with the AdamW optimizer \cite{loshchilov2017decoupled} and CosineAnnealingLR learning rate adjuster \cite{loshchilov2016sgdr} while maintaining a stable diffusion. Additionally, we used a 50\% probability of classifier-free guidance to enforce  inpainting and harmonizing module focuses on the details of the subjects. These prompts were generated by captioning the ground truth image with BLIP2 \cite{li2023blip}.

\vspace{-0.2em}

\section{Experiment}
In this section, we first validate the efficacy of our PhD framework on both  subject-driven editing and generation tasks by comparing it with other baselines.
Then, we perform ablation studies on important design choices, such as the inpainting and harmonizing module and data augmentation training, to understand thier effect.
We also present qualitative examples and include a discussion on observations.

\begin{figure*}[!htb]
    \centering
    \includegraphics[width=\textwidth]{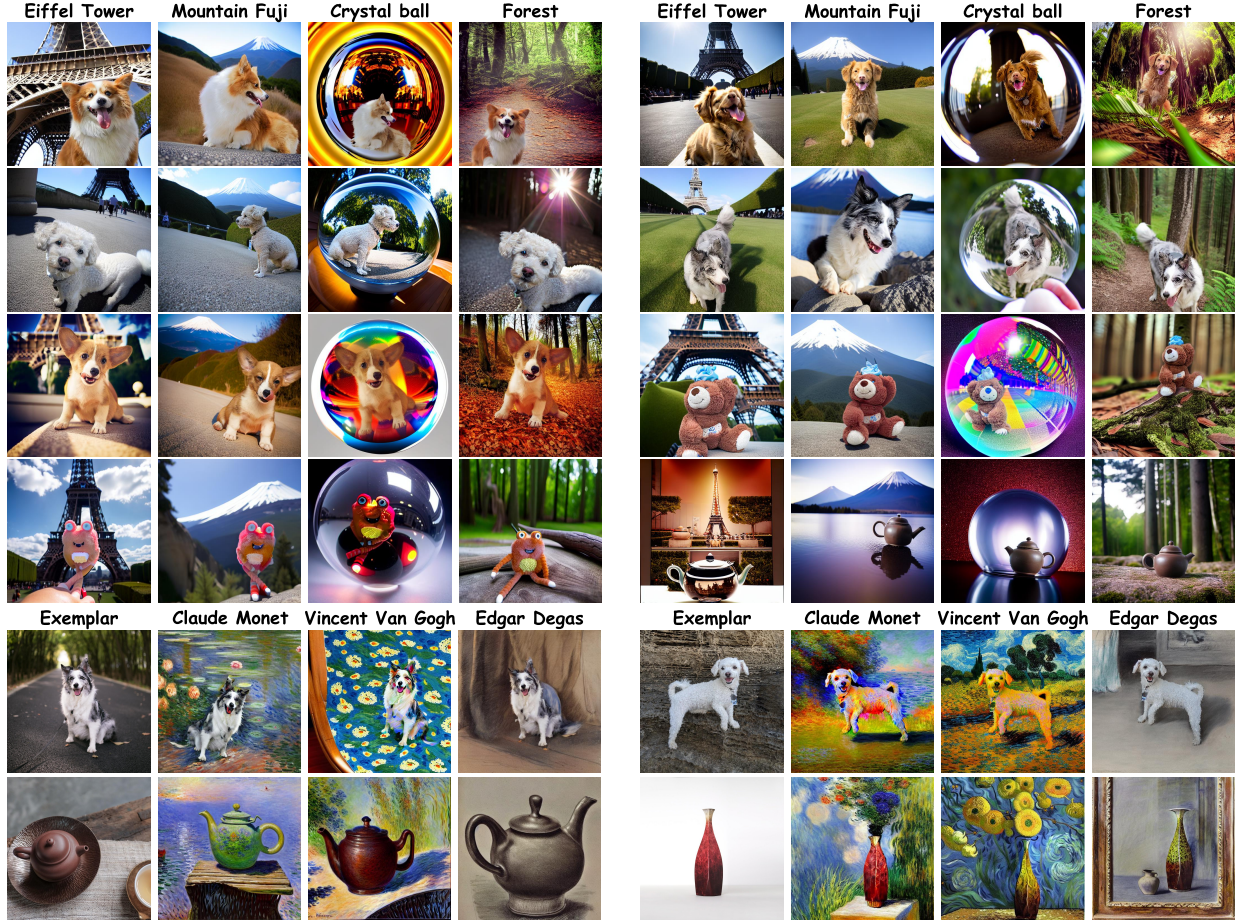}
    \caption{Qualitative results of subject-driven scene generation and style transfer with texts, where the title denotes the name of scene and style.}
    \vspace{-1em}
    \label{fig:scene}
\end{figure*}

\subsection{Subject-Driven Image Editing}
\vspace{-0.2em}
\subsubsection{Test Benchmark}
Following Paint-by-Example \cite{yang2022paint}, we evaluate our method on COCO Exemplar-based Image Editing Benchmark. It comprises 3500 source images from the MSCOCO validation set \cite{lin2014microsoft}. Given a scene image with a corresponding mask and a reference image, the model is tasked with seamlessly blending the subject from the exemplar image into the scene image. To ensure a reasonable combination, the original scene image shares similar semantics with the reference image.
\vspace{-0.2em}
\subsubsection{Competing Methods}
We select recently published state-of-the-art subject-driven image editing methods, Blended diffusion \cite{avrahami2022blended}, DCCF \cite{xue2022dccf} and paint-by-example \cite{yang2022paint}, as baselines. Specifically, Blended Diffusion utilizes a text prompt "a photo of C" to generate the target image, where C represents the subject's class.  In terms of DCCF \cite{xue2022dccf}, we first use LAMA \cite{suvorov2022resolution}  to inpaint the masked area, then paste the inpainted region into the scene image, and finally use DCCF to harmonize the result. In the case of \cite{yang2022paint}, we follow its original implementation, taking the subject image, mask information, and scene image as the model's inputs.

\begin{table}[h]
\centering
\setlength{\tabcolsep}{1.2mm}
\caption{Quantitative comparisons of subject-driven image editing task.
}
\begin{tabular}{cccccc}
\hline
Method & CLIP\textsubscript{I} $\uparrow$ & CLIP\textsubscript{T} $\uparrow$ & FID\textsubscript{scene} $\downarrow$ & FID\textsubscript{ref} $\downarrow$& FID\textsubscript{coco} $\downarrow$\\
\hline
Blended \cite{avrahami2022blended}  & 72.19 & 30.24 & 3.169 & 12.600 & 5.023 \\
DCCF \cite{xue2022dccf} & 76.68 & 29.34 & 2.147 & \textbf{11.206} & 4.354 \\
PbE~\cite{yang2022paint} &76.69 & 29.65 & 2.074 & 12.671 & 4.069\\
Ours & \textbf{78.32} & \textbf{30.53} & \textbf{1.711} & 12.285 & \textbf{4.017} \\

\hline
\end{tabular}
\label{table:clip_score}
\end{table}

\begin{figure*}[!htb]
    \centering
    \includegraphics[width=\textwidth]{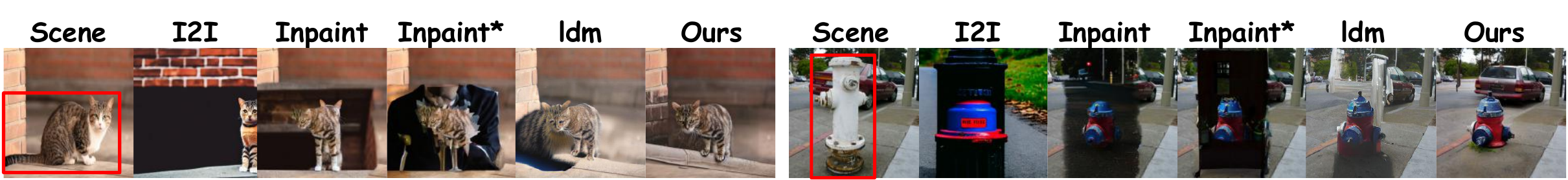}
    \caption{Ablation Study.
    Compare with the naive Stable Diffusion approach.
    I2I denotes the image-to-image pipeline with the edited image $\hat{I}_p$ as input.
    Inpaint denotes inpainting $\hat{I}_p$, Inpaint* is Inpaint with a null prompt, and ldm refers to directly fine-tuning the latent diffusion model. 
    }
\vspace{-1em}
    \label{fig:pipeline}
\end{figure*}

\begin{figure}[!htb]
    \centering    \includegraphics[width=0.7\textwidth]{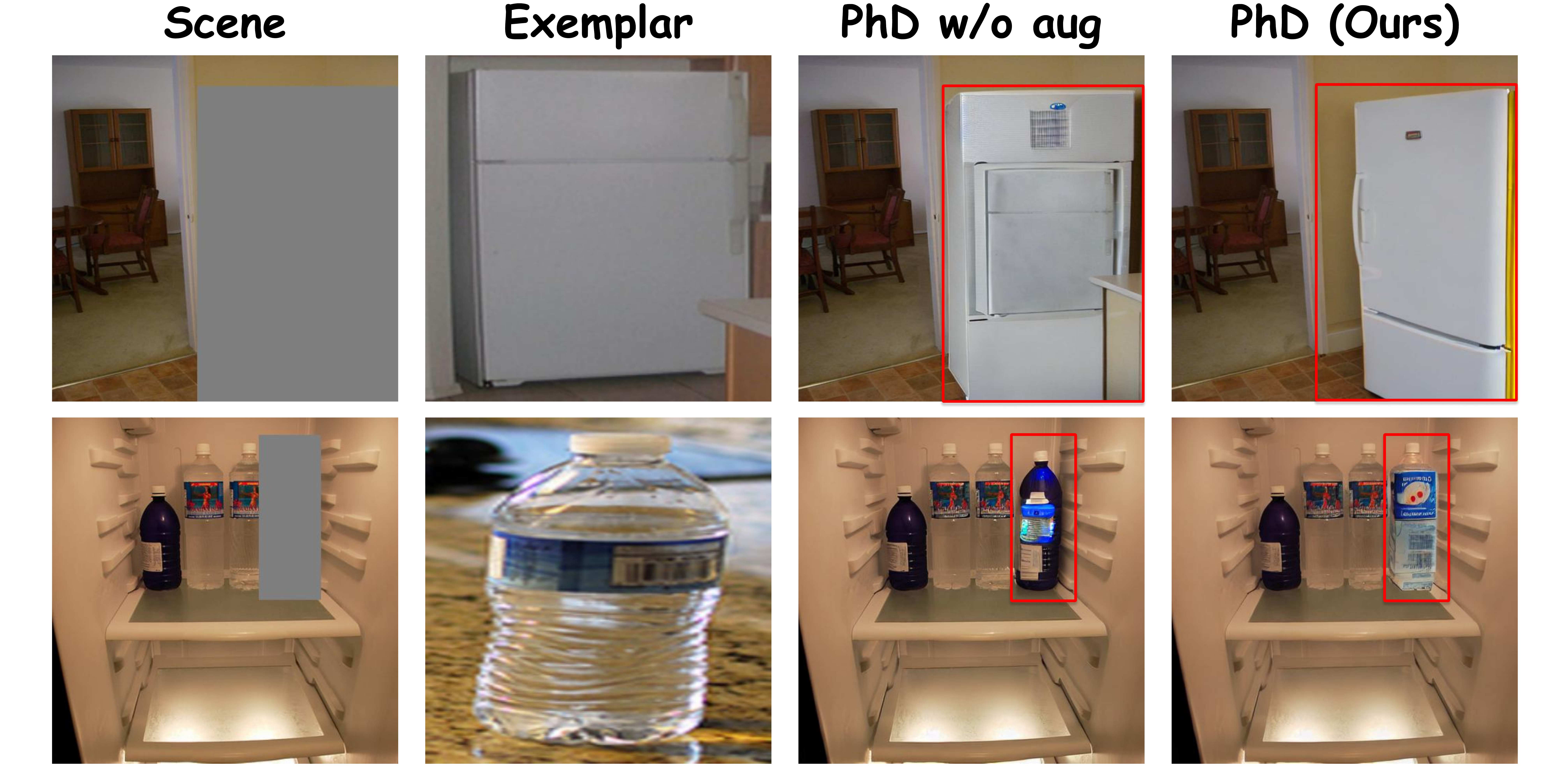}
    \caption{Ablation Study.
    The augmentations help PhD to adapt the exemplar to the scene in a more nature way.
    }
\vspace{-1.2em}
    \label{fig:aba_aug}
\end{figure}
\vspace{-0.2em}
\subsubsection{Quantitative Evaluation}
Subject-driven image editing focuses on seamlessly integrating a user-specific subject into a given scene image. To ensure both image quality and subject identity preservation, we evaluate this task using the following metrics. 1) CLIP\textsubscript{I} computes the CLIP score \cite{radford2021learning} between the exemplar image and the edited image to assess subject identity preservation. 2) CLIP\textsubscript{T} computes the CLIP score between the text captions of the original image from BLIP-2~\cite{li2023blip} and the edited image to evaluate the preservation of semantics. 3) FID\textsubscript{scene} evaluates background preservation quality by computing the FID score \cite{heusel2017gans} between the edited image and the original scene images. 4) FID\textsubscript{ref} assesses subject identity preservation by computing the FID score between the edited image and the exemplar image. 5) FID\textsubscript{coco} metric evaluates overall image quality by calculating the FID score between the edited image and the entire COCO image dataset.

The main quantitative results are shown in Table \ref{table:clip_score}. We can see that our PhD surpasses PbE, the best subject-driven image editing method, by a significant margin (1.71 \emph{versus} 2.07 on FID\textsubscript{scene}), thereby establishing a new state-of-the-art. In addition, DCCF achieves the best FID\textsubscript{ref} because the subject in images generated from DCCF is always the same as the exemplar image but exhibits inconsistencies with the background, resulting in less realistic images, as indicated by other evaluation metrics. In comparison, our method excels in almost all metrics while preserving subject identity, achieving the second-best performance in FID\textsubscript{ref}.

\vspace{-0.2em}
\subsubsection{Qualitative Evaluation}
We provide the qualitative comparisons as Figure \ref{fig:quali_edit} shows, and we can see that Blended Diffusion struggles to maintain the identity of user-specific subjects within the masked area. On the other hand, DCCF effectively preserves the subject's identity, but it often fails to seamlessly integrate the subject into the scene image, leading to results that lack photorealism. These observations are consistent with our quantitative analysis. Paint-by-Example (PbE) employs CLIP to extract the semantic features of subjects, enabling it to successfully retain high-level semantic information. However, the limited expressiveness of CLIP may cause the edited image to lose important low-level details. For instance, PbE can successfully transfer a TV subject from an exemplar image to a scene image, but it may struggle to preserve the content displayed on the TV in the edited image. In contrast, because our method directly paste the subject into the scene, we can effectively preserve low-level details of the subject. In addition, benefited from our inpainting and harmonizing model trained on the augmented subjects, our method can also preserve the semantics of the subject, and even complete the partial subjects to match the contexts in the scene, \emph{e.g.}, PhD is able to complete the human body and vase in the 2nd row of Figure \ref{fig:front-page} and 4th row of Figure \ref{fig:quali_edit}, while other methods cannot. The results clearly demonstrate the superiority of our approach over existing techniques.  
\vspace{-0.2em}
\subsubsection{User Study}

\begin{table}[h]
\centering
\setlength{\tabcolsep}{1.2mm}
\caption{The ratio of the best-selected image among four images. 
"Quality" represents the overall quality of the image.
"Similarity" refers to how closely a user's image matches an exemplar image. 
Users rated ours as the best quality.
}
\begin{tabular}{cccc}
\hline
Method & Quality $\uparrow$ & Similarity $\uparrow$   & Average $\uparrow$ \\
\hline
Blended \cite{avrahami2022blended} & 15.5\% & 11.4\% & 13.5\% \\
DCCF \cite{xue2022dccf} & 13.9\% & \textbf{39.1\%} & 26.5\% \\
PbE~\cite{yang2022paint} & 21.6\% & 10.6\% & 16.1\% \\
Ours & \textbf{48.9\%} & 38.9\% & \textbf{43.9\%} \\

\hline
\end{tabular}
\label{table:user}
\vspace{-0.8em}
\end{table}

Because the quantitative metrics may be not perfect for image synthesis tasks, we carried out a user study involving 25 participants. The study consists of 30 sets of images, with each set containing the scene image, exemplar images and four images output from editing models. The images in each set are displayed in a random sequence for participants. The participants are required to  select the best image for two separate aspects: the overall quality of the edited image and the similarity of the subjects in the edited image with it in the exemplar image. As Table \ref{table:user} shows, the proportion of participants who chose the results edited by our method as having the best quality is significantly higher than that of the baselines. (48.9\% verse 21.6\%). Because ours and DCCF directly copy the subject from the exemplar image, the ratios of our method and DCCF are comparable in terms of subject consistency. However, our method exhibits significantly higher quality in the edited images.

\vspace{-0.2em}
\subsection{Scene Generation and Style Transfer}
Since we keep the parameters of the pre-trained diffusion model frozen and do not replace the textual embedding as in other work \cite{yang2022paint}, we are able to perform subject-driven scene generation and style transfer via textual guidance given a subject.
To demonstrate our method's capability for subject-driven scene generation with textual control, we first extract subject information from datasets proposed by DreamBooth \cite{ruiz2022dreambooth}. We then paste the subject onto a black scene image and designate the entire mask image as the editing area. Using these inputs, our framework can perform scene generation tasks prompted by "a photo of <subject-name> in <scene-name>, in <style-name>." We display some generation samples in Figure \ref{fig:scene}, demonstrating that our method, PhD, can generate high-quality and diverse scenes with the given subject based on textual prompts. We also conduct a quantitative comparison of performance with DreamBooth. The results shown in Table~\ref{table:com_dreambooth}, indicate that PhD's performance is comparable to, and in some cases surpasses, DreamBooth, thus highlighting our framework's ability to preserve the strong scene generation capabilities of the pre-trained diffusion model. The $FID_{ref}$ of Dreambooth is higher than ours because Dreambooth is directly fine-tuned on the reference dataset while ours does not.

\begin{table}[h]
\vspace{-0.5em}
\centering
\setlength{\tabcolsep}{1.2mm}
\caption{Quantitative evaluation metrics on subject-driven scene generation with textual control.}
\begin{tabular}{cccc}
\hline
Method & CLIP\textsubscript{T} $\uparrow$ &FID\textsubscript{ref} $\downarrow$& FID\textsubscript{coco} $\downarrow$\\
\hline
Dreambooth \cite{ruiz2022dreambooth} & 30.00 & \textbf{20.48} & 40.64\\
Ours & \textbf{32.00} & 28.96&  \textbf{35.55} \\

\hline
\end{tabular}
\label{table:com_dreambooth}
\vspace{-1em}
\end{table}

\subsection{Ablation}\label{sec:aba}
\subsubsection{The effects of inpainting and harmonizing model} In order to evaluate the necessity of our inpainting and harmonizing module in our framework, we conduct following ablation studies: 1) In-painting Stable Diffusion (SD), 2)  The I2I SD, and 3) Fine-tuned pre-trained SD on our training datasets. From the results in Figure \ref{fig:pipeline}, we can see that I2I and in-painting SD fail to produce satisfactory results and even cannot preserve the identity of the instance. In addition, we can see that even if SD is fine-tuned on the subject-driven image-editing datasets, it still fails to preserve the instance identity. In contrast, our PhD with inpainting and harmonizing module effectively learns how to blend the pasted subject in the scene image and generate context consistent image.
\vspace{-0.2em}
\subsubsection{The effects of data augmentation} In order to show the effect of the data augmentation strategy in our framework. We show qualitative comparisons between the regular PhD and a version without data augmentation training strategy in Figure \ref{fig:aba_aug}. We can see that the model without data augmentation training strategy normally cannot naturally blend the generated content with the original image, resulting in artifacts and inconsistencies in output. In contrast, the model trained with the data augmentation strategy demonstrates improved performance, with a more seamless integration of the generated content and the scene image. This is because data augmentation strategy helps the model to generalize better by exposing it to a range of input variations during training.

\vspace{-0.2em}
\section{Conclusion}
In this paper, we present a novel framework called Paste and Harmonize via Denoising (PhD) for subject-driven image editing and generation tasks. The framework consists of two steps: the \emph{Paste} step and the \emph{Harmonize via Denoising} step. The \emph{Paste} step allows for the extraction of instance information and its incorporation into the editing image, while the inpainting and harmonizing module in the \emph{Harmonize via Denoising} step guides the pre-trained diffusion model to produce realistic editing results. Our experiments demonstrate that our framework outperforms baseline methods and achieves superior generation results on subject-driven image editing tasks, and show promising results in subject-driven scene generation and style transfer with textual controls.
\vspace{-0.2em}
\paragraph{Limitation and Future Work.}
By applying diverse data augmentation strategies, our PhD research effectively harmonizes the subject into the background scene by adjusting brightness and geometry considering the context. However, it struggles with generating consistent unseen regions of the subject, which we believe can be addressed by introducing 3D information into the framework in future work. 

{\small
\bibliographystyle{ieee_fullname}
\bibliography{egbib}
}
\vspace{-0.2em}

\newpage
\section{Appendix}
We promise that we will release codes and pre-trained models after the acceptance.
\subsection{Details of the Inpainting and Harmonizing Module}
\begin{figure*}[!htb]
    \centering
    \vspace{-1.em}
    \includegraphics[width=\textwidth]{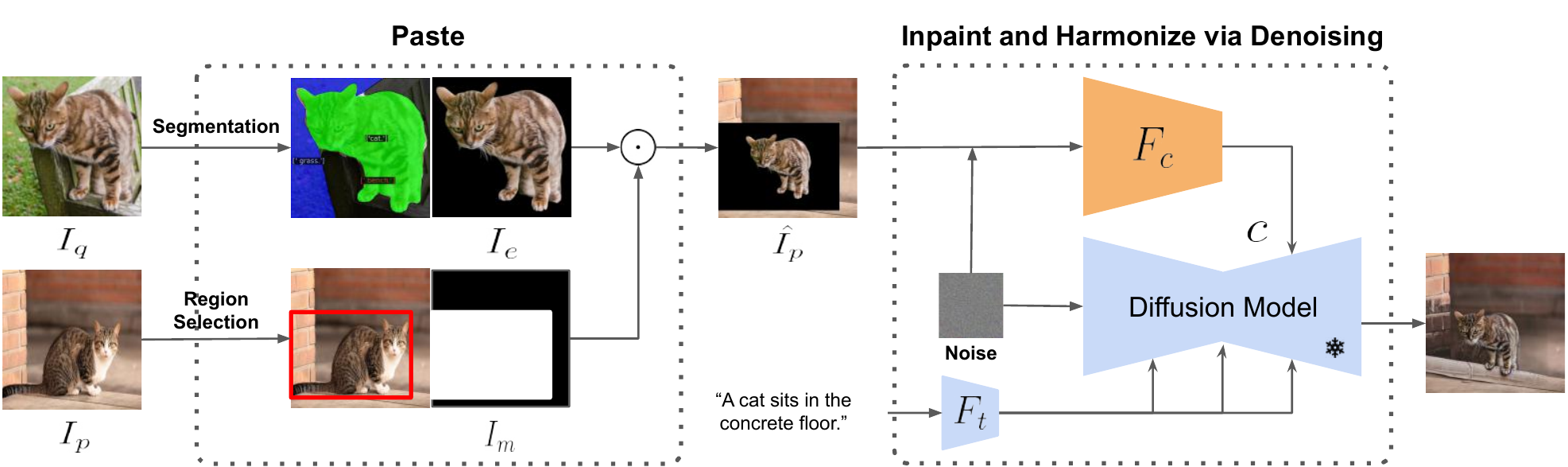}
    \caption{
    The illustration of our proposed Paste, Inpaint and Harmonize via Denoising~(PhD) framework.
    In the Paste step, we extract the subject from exemplar image $I_q$ using the segmentation model and removed the background of the objects within the mask $I_m$ to obtain $I_{e}$. 
    Then we paste $I_{e}$ onto the masked scene image to obtain the pasted image $\hat{I_p}$.
    In the  Inpaint and Harmonize via Denoising step, the Inpainting and Harmonizing module $F_c$ takes $\hat{I_p}$ as the input, and output editing information $c$ to guide the frozen pre-trained diffusion models. The text encoder $F_t$ takes textual prompts as input so that it is able to adjust the style or scene in the edited image.
    }
    \vspace{-1em}
    \label{fig:pipeline}
\end{figure*}
\begin{figure*}[!htb]
    \centering
    \includegraphics[width=\textwidth]{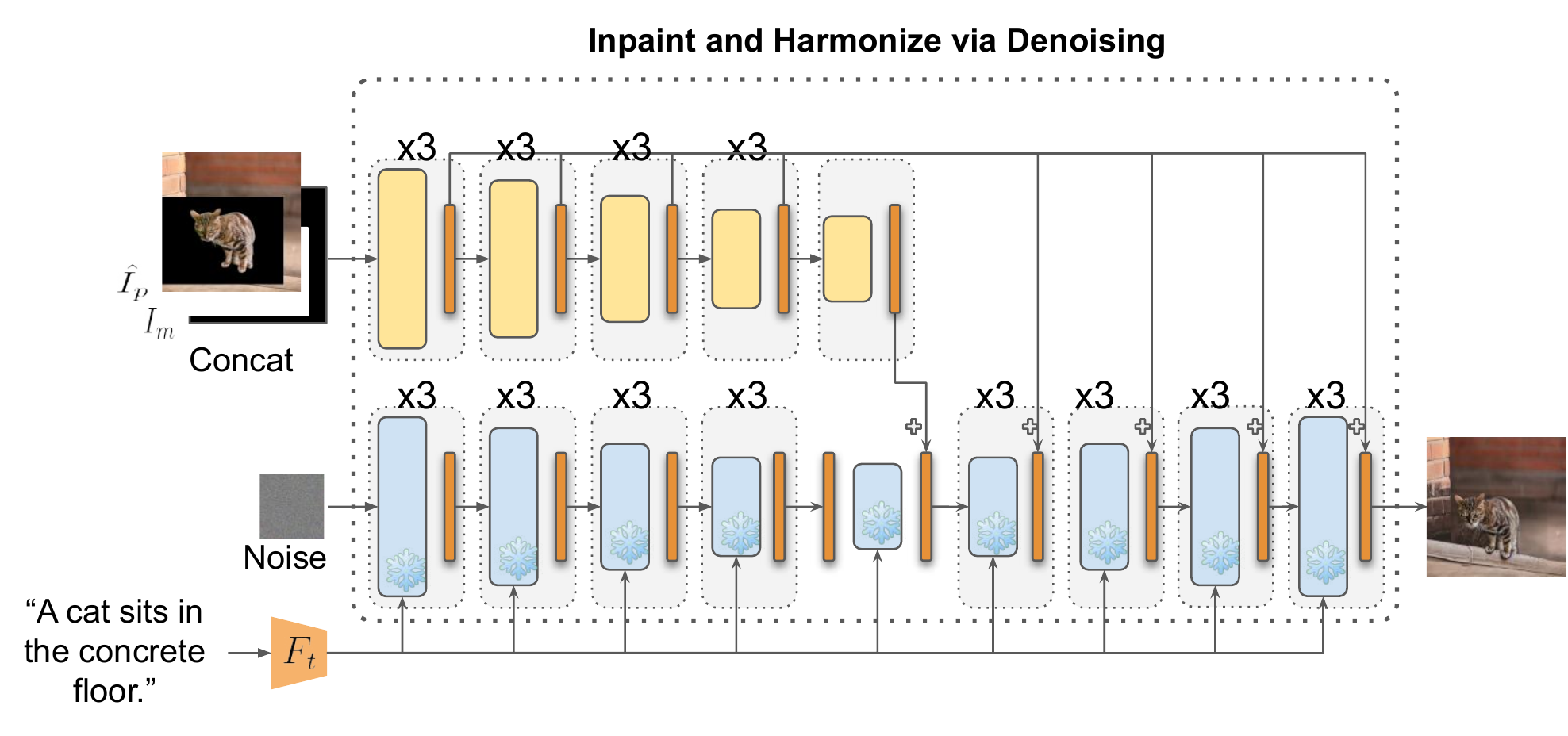}
    \caption{
    The illustration of the Inpainting and Harmonizing module, where the Inpainting and Harmonizing module takes $\hat{I_p}$ and its mask as the input, and output editing information to guide the frozen pre-trained diffusion models. The text encoder $F_t$ takes textual prompts as input and perform cross-attention with the pre-trained U-Net.
    }
    \vspace{-1em}
    \label{fig:inpaint}
\end{figure*}

The detailed illustration is presented in Figure \ref{fig:inpaint}. As can be observed, the inpainting and harmonizing module processes the edited image $\hat{I}_p$ as input and is trained to encode this information to guide the frozen pre-trained diffusion model in blending the subject with the input image. The structure of the inpainting and harmonizing module is inspired by the recent work \cite{zhang2023adding}, employing the encoder architecture of the pre-trained U-Net. Specifically, the feature map extracted by each U-Net block within the inpainting and harmonizing module is added to the corresponding feature map in the U-Net's decoder. This approach ensures that the output of the pre-trained diffusion is cognizant of both subject and context information.

\subsection{Analysis of the Inpainting and Harmonizing Module}
\begin{figure*}[!htb]
    \centering
    \includegraphics[width=\textwidth]{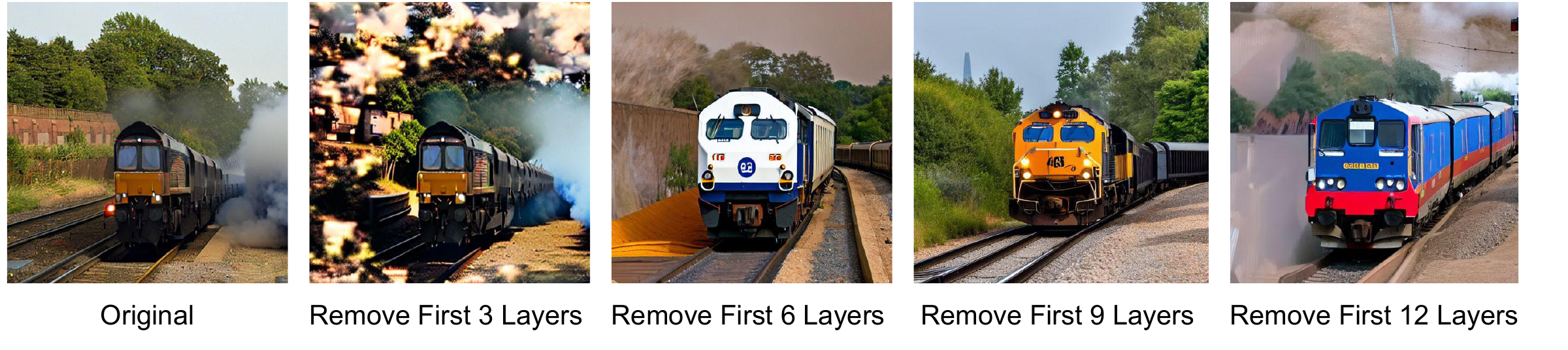}
    \caption{
    The illustration of the Ablation Study of Inpainting and Harmonizing module, where synthesis edited image by selectively removing feature connections in some layers between the Inpainting and Harmonizing module and the pre-trained model.
    }
    \vspace{-1em}
    \label{fig:abl}
\end{figure*}

In order to analyse the Inpainting and Harmonizing Module (IPM), we tried to disconnect the feature connections between Inpainting and Harmonizing Module and the pre-trained diffusion model, and synthesis some edited images. As Figure \ref{fig:abl} shows, the first 3 layers of IPM preserve the background information of the edited image, the first 6 layers of IPM preserve the object information of the edited image, the first 9 layers of IPM preserve the  geometry information  of the edited image, and  the first 12 layers of  IPM preserve the  geometry semantic  information of the edited image.

\begin{figure*}[!htb]
    \centering
    \includegraphics[width=0.96\textwidth]{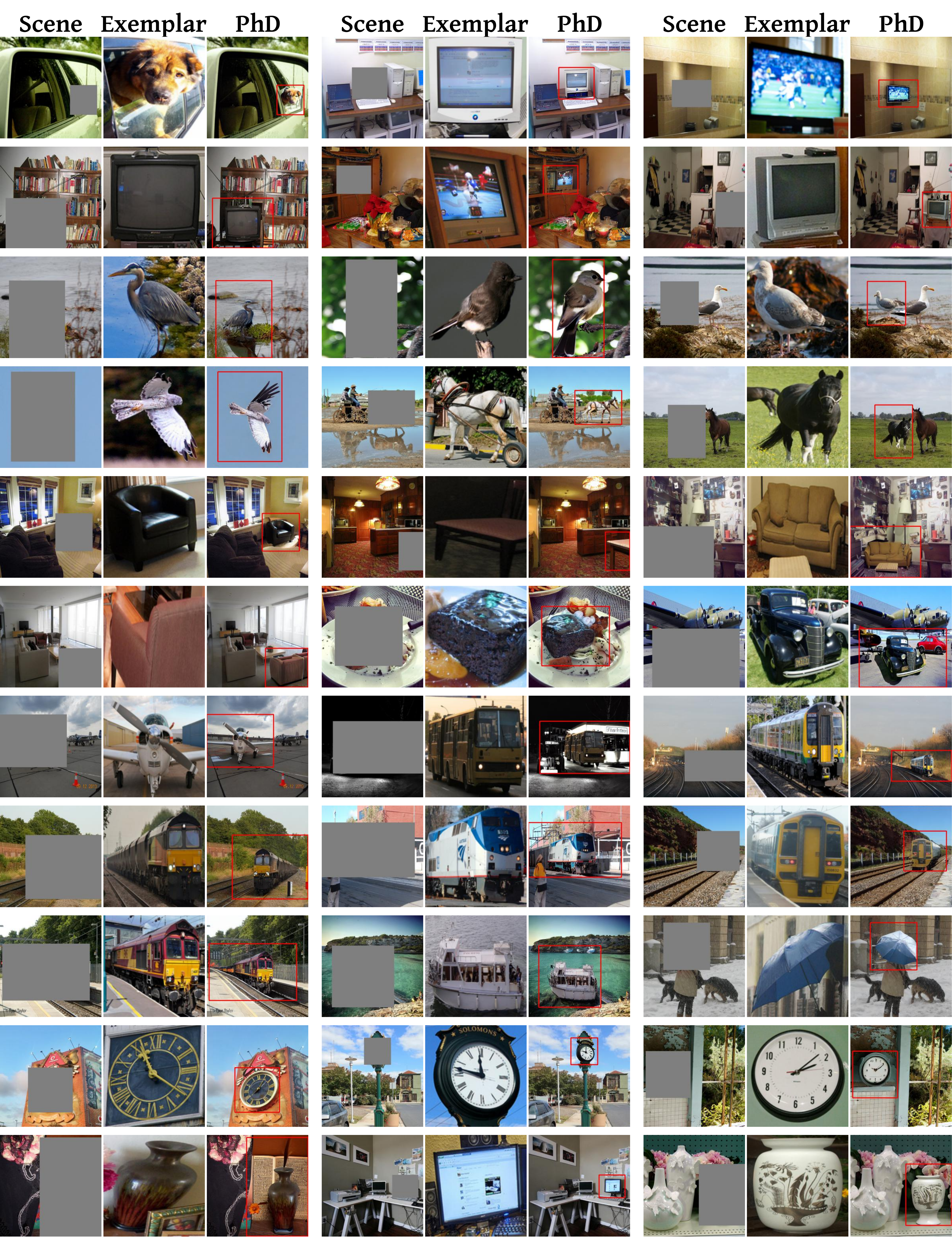}
    \caption{
    Qualitative result of subject-driven image editing tasks.
    }
    \vspace{-1em}
    \label{fig:pipeline}
\end{figure*}

\begin{figure*}[!htb]
    \centering
    \includegraphics[width=0.96\textwidth]{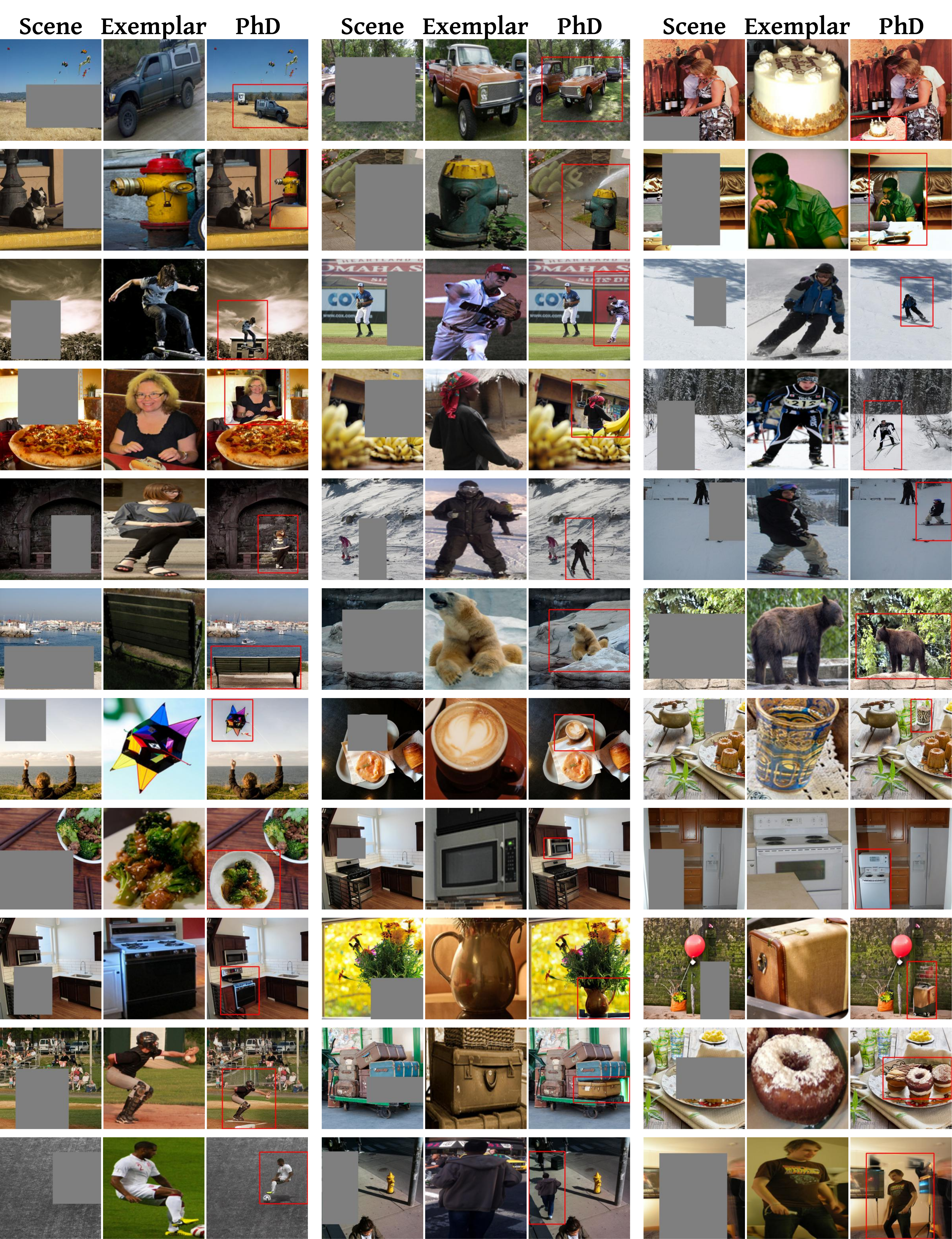}
    \caption{
    Qualitative result of subject-driven image editing tasks.
    }
    \vspace{-1em}
    \label{fig:pipeline}
\end{figure*}


\begin{figure*}[!htb]
    \centering
    \includegraphics[width=0.96\textwidth]{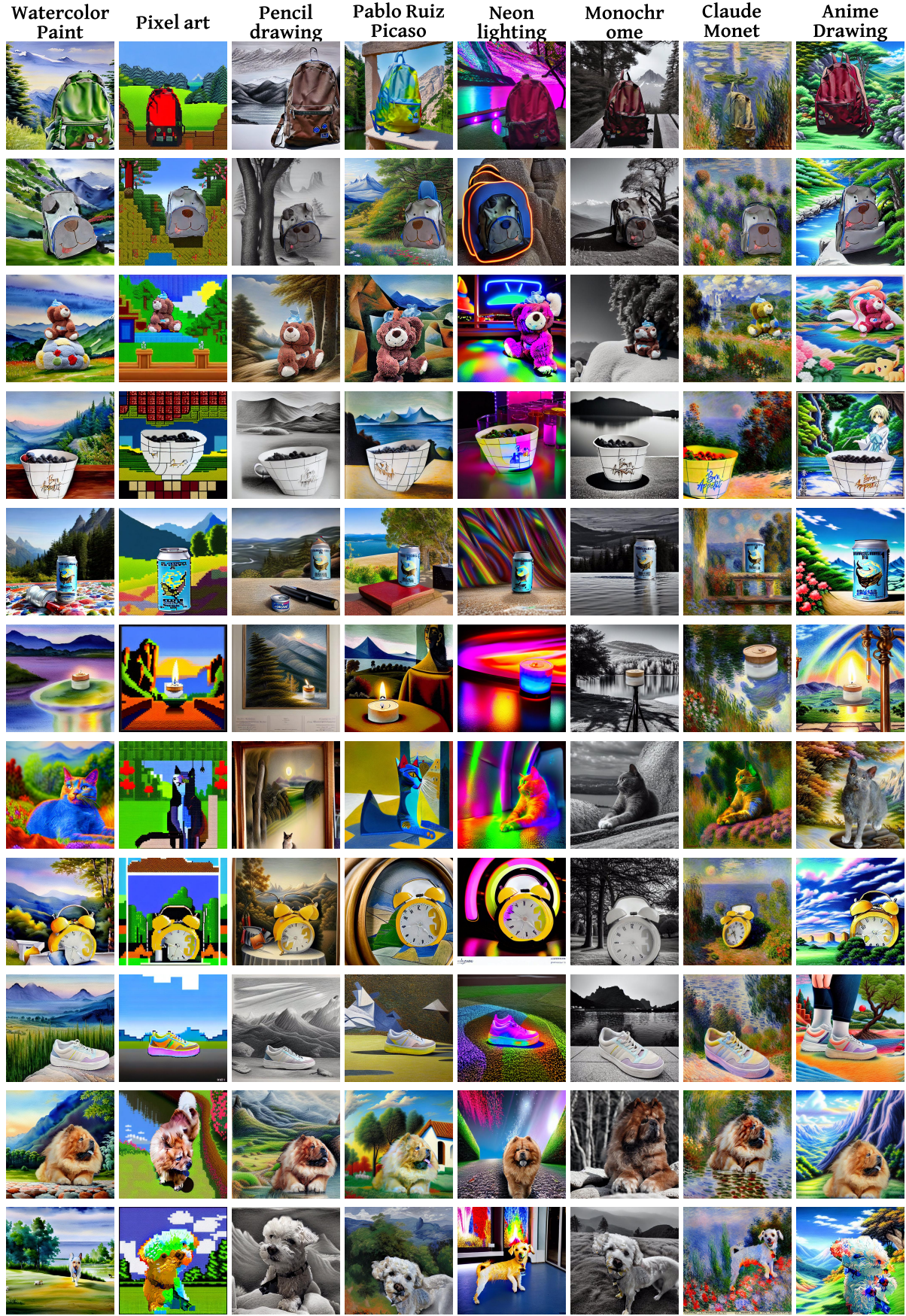}
    \caption{
    Qualitative result of subject-driven scene generation and style transfer with texts.
    }
    \vspace{-1em}
    \label{fig:pipeline}
\end{figure*}

\end{document}